\documentclass{article}

\PassOptionsToPackage{numbers}{natbib}
\usepackage[preprint]{neurips_2021}

\usepackage[utf8]{inputenc} 
\usepackage[T1]{fontenc}    
\usepackage{hyperref}       
\usepackage{url}            
\usepackage{booktabs}       
\usepackage{amsfonts}       
\usepackage{nicefrac}       
\usepackage{microtype}      
\usepackage{xcolor}         

\usepackage{amsmath,amsfonts,bm}

\def\eqref#1{equation~\ref{#1}}

\def\1{\bm{1}}

\DeclareMathAlphabet{\mathsfit}{\encodingdefault}{\sfdefault}{m}{sl}
\SetMathAlphabet{\mathsfit}{bold}{\encodingdefault}{\sfdefault}{bx}{n}

\usepackage{graphicx}
\usepackage{subfigure}
\usepackage{amsmath}
\usepackage{enumitem}
\usepackage{xcolor}
\usepackage{booktabs}
\usepackage{array}
\usepackage{multirow}
\usepackage{multicol}
\usepackage[numbers]{natbib}

\title{NeuroView: Explainable Deep Network \\ Decision Making}

\author{%
  CJ Barberan \\
  Department of Electrical \\ and Computer Engineering\\
 Rice University\\
  Houston, TX 77005 \\
  \texttt{cb30@rice.edu} 
  \AND Randall Balestriero \\
  Department of Electrical \\ and Computer Engineering\\
 Rice University\\
  Houston, TX 77005 \\
  \texttt{randallbalestriero@gmail.com}

  \AND Richard G. Baraniuk  \\
  
 Department of Electrical \\ and Computer Engineering\\
 Rice University\\
  Houston, TX 77005 \\
  \texttt{richb@rice.edu} 
}

\begin{document}

\maketitle

\begin{abstract}
Deep neural networks (DNs) provide superhuman performance in numerous computer vision tasks, yet it remains unclear exactly which of a DN's units contribute to a particular decision. 
{\em NeuroView} is a new family of DN architectures that are interpretable/explainable by design. 
Each member of the family is derived from a standard DN architecture by vector quantizing the unit output values and feeding them into a global linear classifier.  
The resulting architecture establishes a direct, causal link between the state of each unit and the classification decision.
We validate NeuroView on standard datasets and classification tasks to show that how its unit/class mapping aids in understanding the decision-making process. 
\end{abstract}

\section{Introduction}
Deep networks (DN) have become the de facto approach in numerous machine learning problems. 
However, by and large, they remain opaque black boxes whose decisions can be challenging to interpret. One example is the colored MNIST dataset \cite{kim2019learning}, where typical classifiers align the prediction based on the color instead of the shape/contour of the number.
It would be important to explain which of the units are responsible for classifying the digit and \textit{how much} the units are contributing to the decision as opposed to inferring from results-based analysis. 
Interpretability is essential in many fields to figure out if there are any biases from the DN. For example, in \cite{nam2020learning}, the authors performed experiments on an action recognition dataset where the test accuracy is far from stellar due to the action recognition biases.


Interpretability has been defined as ``comprehending what the model has done'' \cite{gilpin2018explaining}. The authors  in \cite{gilpin2018explaining} also stated that the main goal of interpretability is to detail what the model's internal workings are in a way that is understandable to a human.
There have been works dealing with single-unit interpretability using visual concepts  \cite{bau2017network, bau2018gan, bau2020understanding}. There have been other works, \cite{bau2020understanding,ghorbani2020neuron} denoting how removing a unit affects the classification accuracy. However, the main issue with these works is that they cannot provide explanations. In \cite{gilpin2018explaining}, the authors state that it is often difficult to have a network that is easily interpretable to humans and explainable.

There are other interpretable works that delve into inspecting which portions of the input images is the network focusing. One other line of work are the saliency methods \cite{selvaraju2017grad,shrikumar2017learning,sundararajan2017axiomatic} while another line of work is the perturbation methods \cite{ribeiro2016model,fong2017interpretable,singh2018hierarchical}. Their use case is on an image-to-image basis but do not provide much information on the class level. There are other works in wanting to use interpretability and explainability like \cite{kim2018interpretability,ghorbani2019towards,goyal2019explaining} where they explain the classifiers using human-interpretable visual concepts. However this analysis is done on a layer-to-layer basis that cannot tell anyone how this connects within the class level.

This prior work has focused on single unit, layer-wise, or input image interpretability for DNs but one key question that is the focus of this paper, namely \textit{how all the units within the network coordinate to explain the decision-making process}, needs more attention. Interpretability is not enough and there needs to be more aspects of explainability. 

NeuroView modifies existing DNs by providing a direct connection between each unit in every layer to the output, thereby making every unit visible. This is in stark contrast to traditional DNs where all layers except the last layer are hidden. Hence with modification, we can inspect the linear classifier's weight to explain which units are contributing the most per class.

\begin{figure}[t!]
    \centering
    \begin{tabular}{cccc}
        \includegraphics[width=0.22\textwidth]{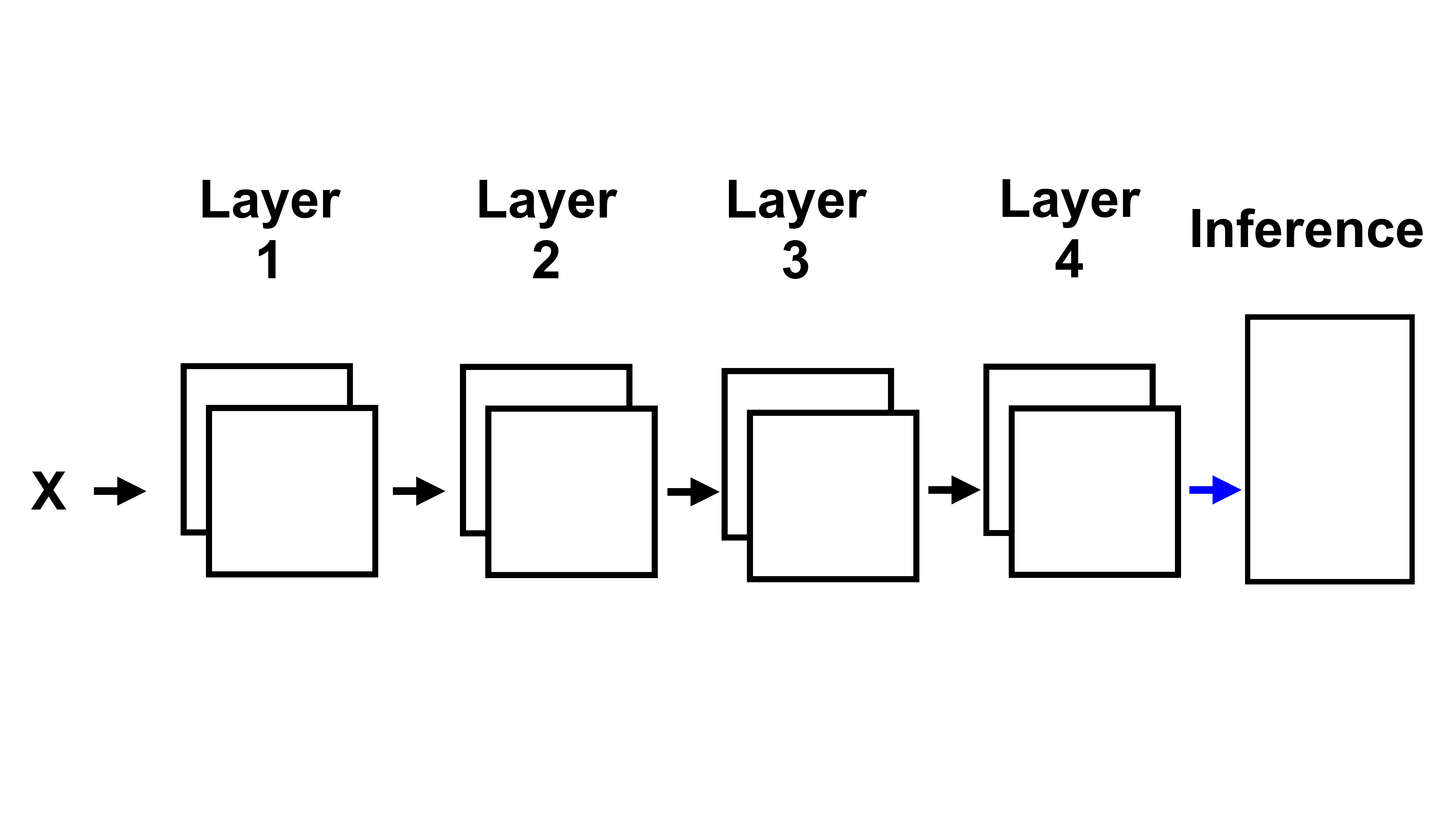} & 
        \includegraphics[width=0.22\textwidth]{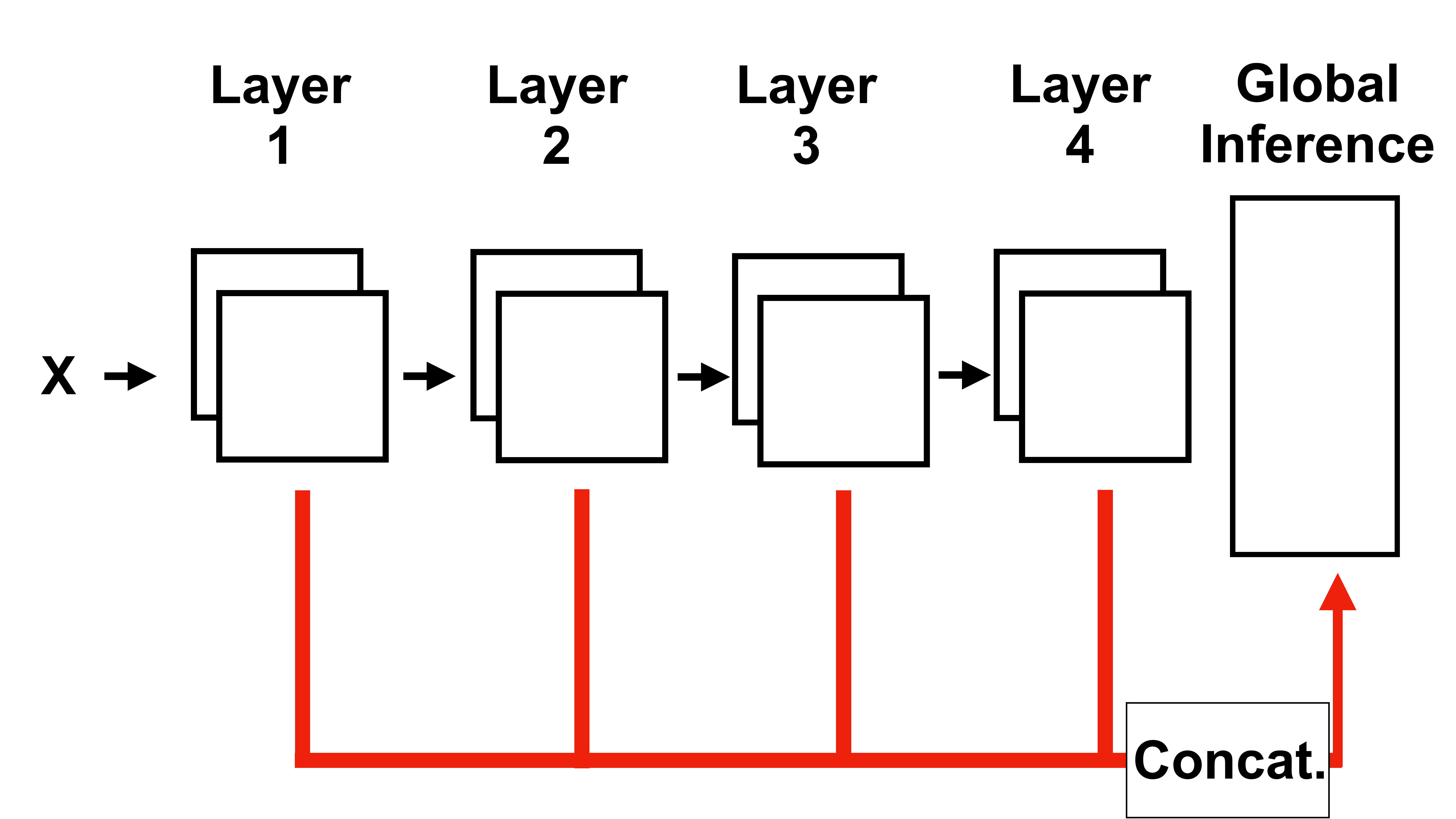} &\includegraphics[width=0.22\textwidth]{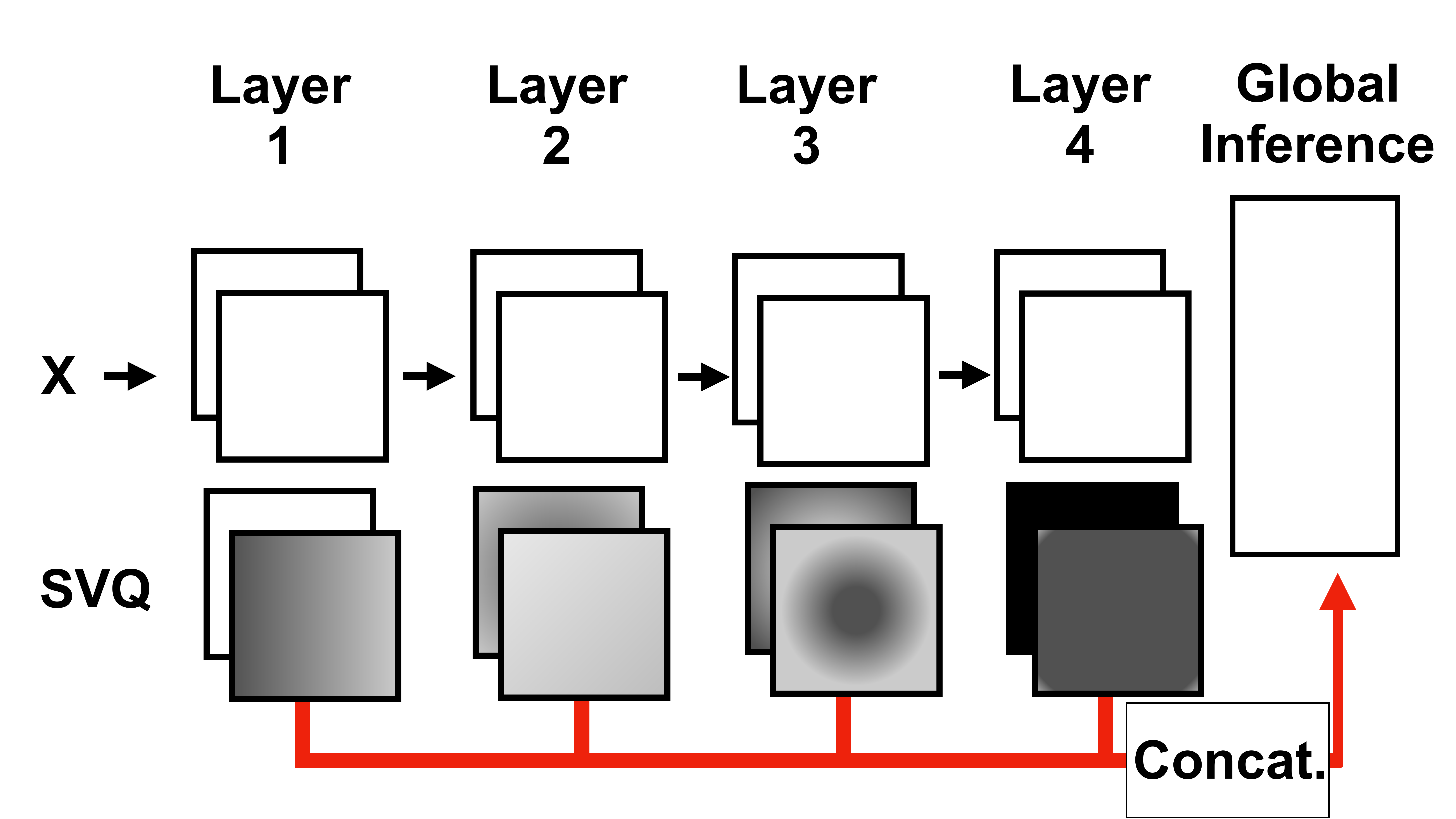} &
        \includegraphics[width=0.22\textwidth]{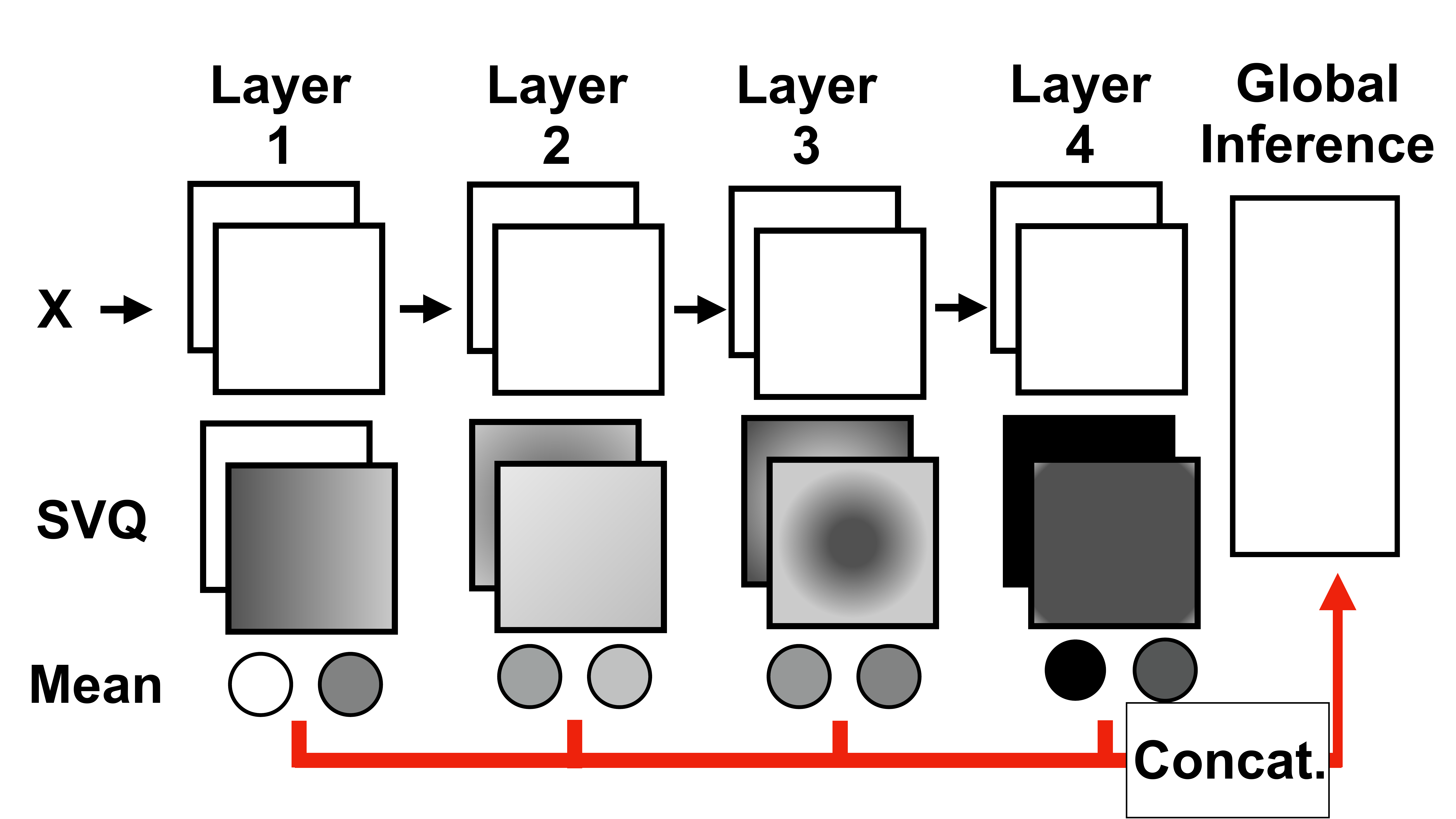}\\
        (a) & (b)  & (c) & (d)\\
    \end{tabular}
    \caption{\small Schematic of the conversion of a deep network (a) to its NeuroView variant (b). (c) The Soft VQ code for NeuroView.
     The Soft VQ is denoted as a sigmoid function applied to the layer.
    Black denotes 1 while white denotes 0. Gray colors in between denote values between 0 and 1. (d) Max takes the maximum value for each unit while mean takes the average value for each unit as the input to the inference layer.
    }
    \label{fig:NVVariant}
\end{figure}

{\bf The NeuroView Family of DN Archiectures.}~
NeuroView is a new family of 
DN architectures that can explain which units are responsible for classification in a quantitative manner.
Each member of the family is derived from a standard (difficult to interpret) DN architecture (e.g., Convolutional Neural Network (CNN), ResNet \cite{he2016deep}, DenseNet \cite{huang2017densely}) via the following procedure (see Fig.~\ref{fig:NVVariant} for a schematic):
\begin{enumerate}[leftmargin=*]
\setlength\itemsep{0em}

\item
{\bf To enable each DN unit to directly contribute to classification}, 
we remove the final inference layer of the DN (e.g., linear classifier and softmax in a classification net) and instead feed each unit's output directly into one (very wide) linear classifier.
\item
{\bf To enhance the interpretability of each unit's contribution to classification}, we use the DN's activations.
We also demonstrate how those activations can be made differentiable by applying a Vector Quantization (VQ) process.

\item
{\bf Explain the classification by inspecting which units contribute the most for that class}.

\end{enumerate}

\begin{figure}[t!]
    \centering
    
    \includegraphics[width=100mm]{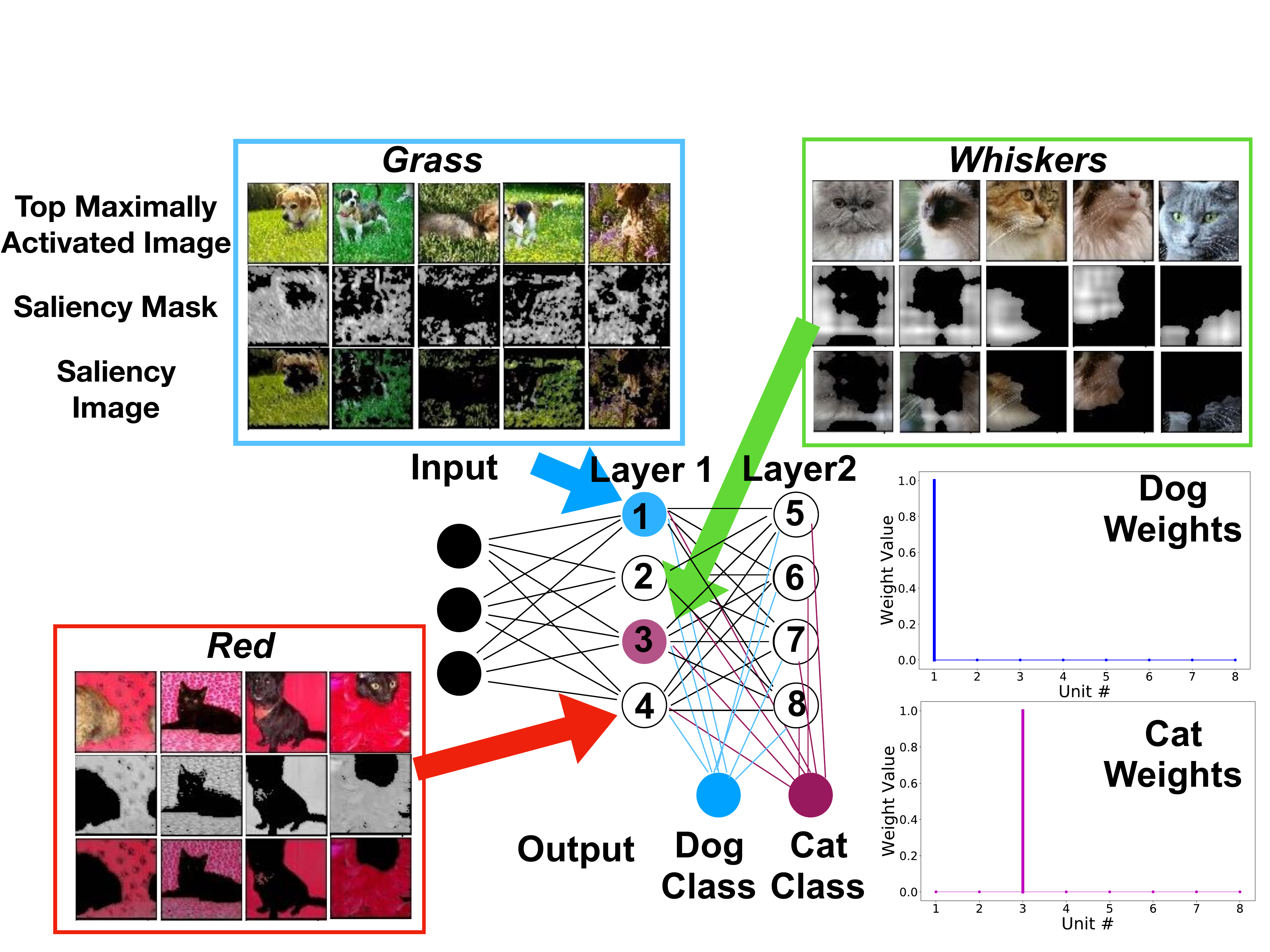} 
    \caption{\small A representation of the NeuroView network. All of the units are connected to the linear classifier that allows easy explainability to which units are contributing to the decision making. For the dog class, the first unit contributes the most while for the cat class, the third unit contributes the most. Then by using other interpretability methods, this provides additional understanding of what each unit is learning. 
    }
    \label{fig:NVCore}
\end{figure}

We summarize our contributions as follows:

{\bf [C1]}~We demonstrate empirically with a range of architectures and datasets that the NeuroView network has on par accuracy performance to its DN equivalent (Table~\ref{tab:my_label} in Section 3).

{\bf [C2]}~We show quantitatively which of the units within the NeuroView network are contributing to the decision-making process of the classification task (Fig.~\ref{fig:NVCore}) by inspecting the weights of NeuroView's linear classifier. 

{\bf [C3]}~We add an additional layer of interpretability by using Network Dissection \cite{bau2017network} to acquire the human-interpretable concepts, like color, textures, objects, and scenes, for the NeuroView units. We display which concepts are prevalent for every class in Fig.~\ref{fig:ClassConcepts} by displaying the class concept distribution.

{\bf [C4]}~We show multiple examples to illustrate how useful NeuroView can be in providing more information about the inner machinations of the network. We also demonstrate how NeuroView can provide additional understanding in a range of tasks from ordinary object classification, multi-view object classification, stylization, and sound classification.


\section{Background}
To understand NeuroView, we first dive into neural networks, and representing them by their respective operators.

{\bf Deep Networks.}~
A DN is an operator $f$ that maps an input $x \in \mathbb R^n$ to an output $y \in \mathbb R$ (in general) by composing $L$ intermediate {\em layer} mappings $f_{\ell}$, $\ell=1,\dots,L$. Each layer combines affine and nonlinear operators such as the {\em fully connected operator}, {\em convolution operator}, {\em ReLU operator}, or {\em pooling operator}. 
Precise definitions of these operators can be found in \cite{goodfellow2016deep}. We focus on layers made of a composition of an arbitrary number of linear operators preceding a single nonlinear operator. 

Each layer's nonlinearity has a (per-layer) code denoted as $q_{\ell}$ where the code is defined as the derivative of the activation function. Others have used different codes in deep learning like binarization \cite{lin2015deep,courbariaux2016binarized} but we will use the code defined recently.
This code also exists for smooth nonlinearities such as gated linear units \cite{balestriero2018from}, in which case it corresponds to 
\begin{align*}
    [q_{\ell}(f_{\ell-1})]_i &= \frac{e^{[W_{\ell}f_{\ell-1}+b_{\ell}]_i}}{1+e^{[W_{\ell}f_{\ell-1}+b_{\ell}]_i}}\\
    &=\text{sigmoid}([W_{\ell}f_{\ell-1}+b_{\ell}]_i),
\end{align*}
where $W_{\ell}$ and $b_{\ell}$ are the weights and biases of that layer and $f_{\ell-1}$ is the previous feedforward layer mapping.

\section{NeuroView: A Family of Explainable Deep Learning Architectures}

\subsection{Enable each Unit to Directly Contribute to Classification}

 The first step of NeuroView is to gather all the operators of a given DN architecture. This is done simply by acquiring any such nonlinear unit of any layer and concatenating them into a feature vector that we abbreviate as $z$, 
\begin{align*}
   z=
  [f_{1}(x)^T,f_{2}(x)^T,\dots,f_{L}(x)^T]^T
\end{align*}

The concatenation of other layers' activations into the linear classifier has been featured in \cite{lin2017feature}. Yet that work only uses the layers closest to the fully-connected layer which have semantic information that they want to utilize for the fully-connected layer. In addition they use a 1$\times$1 convolution layer to reduce the channel dimensions, where for our work the channel dimensions is vital since we want to see which units from each layer is contributing the most for the linear classifier. In addition, \cite{lin2017feature} focuses on the performance increase in object detection while our work focuses on interpretability/explainability for several 2D convolutional DN tasks.

This concatenated feature vector is then denoted as the {\em view} of the input $x$ from the DN perspective. In order to provide an interpretable classification based on this DN view, we feed the concatenated feature vector into a linear classifier that outputs the class prediction as
\begin{align}
    \widehat{y}=Wz+b,\label{eq:first}
\end{align}
with $\widehat{y}$ the predicted output, $W$ the matrix of linear classifier weights with number of rows depending on the number of classes/output dimensions, and $b$ the classifier bias. For classification, we apply a softmax operation in Equation~\ref{eq:first}. 
The use of a linear classifier is crucial as it is straightforward to interpret the role of each input into the forming of the prediction \cite{kim2007visualizable,james2009functional}.


\subsection{Soft VQ to Enhance the Interpretability of Each Unit/Neuron’s Contribution to the Inference}
We now propose to replace the concatenated feature vector by the associated Soft VQ code from \eqref{eq:first}. The reason is that the feature vector spans all real numbers. The values of te feature vector are converted to a number between 0 and 1 with Soft VQ to easily quanitfy the contribution of each entry. From this perspective, the input to the linear classifier is now a high dimensional vector as 
\begin{align}
    \widehat{y}=\mathrm{softmax}(Wq(z)+b),\label{eq:second}
\end{align}
where $q(z)$ is the Soft VQ of the units that were selected as inputs of the linear classifier (Equation \ref{eq:second}). The Soft VQ depiction is in Fig.~\ref{fig:NVVariant} (c).


\subsection{Dimensionality Reduction of the Soft VQ Codes}

Typical DNs are increasing in size where some have millions of units. This results in their activations being greater than the number of units. This implies Soft VQ codes will be greater than the number of units. Hence the goal is to reduce the Soft VQ code to the length of the number of units of the network. This is achieved by applying an operation that is depicted in Fig.~\ref{fig:NVVariant} (d).


\subsection{Interpretable Linear Classifier Input}
Given in the earlier subsections we fixated on the individual units and focusing on the interpretability of the units. Since we concatenate all of the units in the form of the Soft VQ code. With the Soft VQ code, we can inspect which units contribute from a range of zero to one as the input to the linear classifier. Hence inputs from same class should have similar Soft VQ codes.

\subsection{Explainable Linear Classifier}
When training is complete, we look at the weights, $W$ from \ref{eq:first}, per class
and since we have the linear relationship of the linear classifier to all the units that is represented as the Soft VQ codes, we have this ordered mapping of the weights to the units. This ordering is done by concatenating the Soft VQ codes starting from the early layer's Soft VQ code and concluding with the final layer's Soft VQ code.

Once a DN has been turned into a NeuroView architecture and has been trained to solve the task at hand, we leverage the linear Soft VQ--class mapping for explainability. In training, both the network and linear classifier are trained together and from scratch. With the NeuroView network trained, we focus on the explainability by inspecting the units of the linear classifier and observing the distribution. This Soft VQ combined with the linear classifier tells us how much a unit impacts (linearly) with respect to the outputs of the linear mappings. Based on the amplitudes of the linear classifier matrix weights coming from the units and going to all the classes, we can thus quantify how the units directly amplify or lessen the prediction of the different classes.
 
 Hence, we inspect which of the units are contributing to the decision-making process for every class by inspecting the linear classifier's weights for that class. Figure~\ref{fig:BallroomLinearLayer} shows the weights of the ``ballroom'' class and the beginning index corresponds to the first convolutional layer units while the later index corresponds to the last convolutional layer units. By having our network provide the weights, we can assess quantitatively which units are the most responsible in predicting for each class. Hence we can explain from the weights which units contribute the most per class. From there we can provide additional single unit interpretability techniques to enhance the explainability. In addition, by looking at each class's weights we can observe the different weight distributions. Classes that are similar should have similar weight distributions while classes that are different should vary with how the weights are prioritized by the linear classifier.




{\bf Datasets.}~ We sample 10 classes from the Places365 dataset, which we denote as Places10, \cite{zhou2017places} to assess how NeuroView compares against the DN equivalent since training on all 365 classes takes much longer. In addition, we use the ModelNet40 dataset \cite{wu20153d} for object classification as well as formatting it to do multi-view (MV) object classification \cite{su2015multi}. In MV classification, 12 views are the input as opposed to one view in the regular object classification scheme. In addition, we use a dataset from \cite{huang2020compact} that has 64 classes of different textures. We denote this dataset as Texture. Plus, a non natural image dataset denoted as UrbanSound8k \cite{salamon2014dataset} is used where the input to the network is a spectrogram of the sound clip. This dataset has 10 classes of different sounds.


{\bf Results.}~Table~\ref{tab:my_label} shows that the NeuroView networks will have on par performance with the equivalent DN. For the NeuroView network, we train from scratch. From Table~\ref{tab:my_label} NeuroView is able to be on par or a bit better in terms of accuracy for regular computer vision tasks and even non typical image tasks like the sound classification. NeuroView is able to be on par with DNs and this is important since the authors from \cite{gilpin2018explaining} state that many explainable methods lack the best accuracy.



\begin{table*}[t]
    \centering
    \caption{ \small
    Validation/Test accuracy performance among different datasets. Even in different domains, the NeuroView networks still retain good performance compared with the equivalent DN.}
        {\small 
    \begin{tabular}{llcc} \hline
        \toprule
         Network & Activation & Dataset  & Validation/Testing Accuracy  \\ 
         \midrule
         ResNet18 \cite{he2016deep} & --- & Places10 & 93.7 \\ 
         NeuroView ResNet18 & Max Soft VQ & Places10 & 93.1 \\  \hline 
         ResNet34 \cite{he2016deep} & --- & Places10 & 92.3\\ 
         NeuroView ResNet34 & Max Soft VQ & Places10 & 93.9\\ \hline 
         Resnet50 \cite{he2016deep} & --- & Places10 & 92.7\\ 
         NeuroView Resnet50 & Max Soft VQ & Places10 & 91.9 \\ \hline 
         VGG11 \cite{simonyan2014very} & --- & Places10 &  92.9 \\ 
         NeuroView VGG11 & Max Soft VQ & Places10 & {\bf 94.0} \\  \hline 
         VGG19 \cite{simonyan2014very} & --- & Places10 & 92.9 \\ 
         NeuroView VGG19 & Max Soft VQ & Places10 & 92.8 \\ \hline 
         Alexnet \cite{krizhevsky2012imagenet} & --- & Places10 & 79.6 \\ 
         NeuroView Alexnet & Mean Soft VQ & Places10 & 89.6 \\ 
         NeuroView Stylized Alexnet & Mean Soft VQ & Places10 & 89.3 \\ \hline 
         VGG11 \cite{simonyan2014very} & --- & ModelNet40 & 92.5 \\ 
         NeuroView VGG11 & Mean Soft VQ & ModelNet40 & \textbf{92.7} \\  \hline 
         VGG11 \cite{simonyan2014very} & --- & ModelNet40 MV & 90.9 \\ 
         NeuroView VGG11 & Mean Soft VQ & ModelNet40 MV & \textbf{91.6} \\ \hline 
         VGG11 \cite{simonyan2014very} & --- & Texture & 96.8 \\ 
         NeuroView VGG11 & Mean Soft VQ & Texture & \textbf{99.3} \\  \hline 
         Alexnet \cite{krizhevsky2012imagenet} & --- & UrbanSound8k & 73.1 \\ 
         NeuroView Alexnet & Mean Soft VQ & UrbanSound8k & \textbf{74.3} \\ 
         \bottomrule
    \end{tabular}
    }
    \label{tab:my_label}
    \vspace*{-2mm}
\end{table*}

\section{NeuroView Case Studies}

We demonstrated above how to turn any DN architecture into a NeuroView network and showed that the NeuroView networks' performances are on par with the DN equivalent. This linear Soft VQ of the units--class mapping  is the cornerstone of the following explainable results that enables us to understand more about what is happening inside the machinations of the network. We present various analyses in the forms of different case studies to illustrate how NeuroView can provide additional understanding that is different from results-based analysis.

\subsection{Case Study 1: Unit--Class Linear Mapping}
The issue that is present is that in \cite{bau2017network}, the framework provides which unit aligns the most with the visual concept in their annotated dataset of concepts. The missing component is that we do not know which concepts align with each class for respective task. Since a DN is nonlinear, just knowing the unit's concept cannot tell us how it impacts the decision making. This is where NeuroView can connect the decision making with the concepts. In addition, NeuroView can tell us which concepts negatively impact the class where even \cite{bau2017network} cannot provide which concept negatively impact the unit.

NeuroView provides how each unit impacts the decision making for each class via the linear classifier's weights. Figure~\ref{fig:BallroomLinearLayer} (a) and (c) show the linear classifier's weights for different classes and different networks for Places10. To provide additional interpretability, we leverage \cite{bau2017network} to align the best concept with every unit. Hence, for every class, we sum all the weight values from the linear classifier with their respective concepts and provide a concept--class mapping. Figure~\ref{fig:ClassConcepts} shows the top 5 positive and negative concepts for the ``ballroom'' and ``barn'' class. For ``ballroom'' the dominant positive concept is the chequered texture concept and the dominant negative concept is sky. While the dominant positive concept for ``barn'' is sky and the dominant negative concept for ``barn'' is the grid texture concept. With NeuroView and \cite{bau2017network}'s labels for the units, there is a mapping of the concepts to the classes.

\subsection{Case Study 2: Neural Style Transfer}

Stylization/Neural Style Transfer is a technique from \cite{gatys2015neural} where they take the style of one image and transfer it to a different image. This has been applied from taking the artistic style from one artist like Van Gogh and applying it to a random image. The concern is that the authors in \cite{geirhos2018imagenet} only show a results-based analysis that networks trained with these stylized images have less of a texture bias. This is where NeuroView comes in to understand what is happening within and why the networks have an aversion to textures. From the Places10 dataset, we increase the amount of data by applying 10 random styles to the dataset.

From the Neural Style Transfer experiment, Figure~\ref{fig:BallroomLinearLayer} (d) shows how the early layer weights in a stylized model have higher negative weights than a non stylized model. The main difference is that one is trained with additional stylized images \cite{gatys2015neural}. Either NeuroView network, stylized or not, has similar validation accuracy performance in Table~\ref{tab:my_label}. Even though they have the same performance, their linear classifier's weight prioritization is very different. This prioritization is very evident with the negative weights in the earlier layers. The weights that contribute the most in a negative manner in the stylized NeuroView Alexnet network come from the early layers. From \cite{bau2017network}, the authors state that early layer units align more with colors and textures.



\subsection{Case Study 3: Inner Workings with Different Datasets}
Researchers have used CNNs in different datasets that vary in composition of the data creation. If the network provides the best accuracy through experimental findings, then researchers will adapt them into other datasets. Yet, the issue is without inspecting the inner workings of the network, will the network behave in the same manner? There can be issues with the metric being only accuracy since it is viewed as short-sighted. For instance, \cite{recht2018cifar} showed that CIFAR-10 classifiers could not generalize to the CIFAR-100 dataset. Henceforth, this is where NeuroView provides additional understanding. With the weights of the linear classifier from a NeuroView network, we can inspect the weights and assess if the weights are the same from two datasets.

\begin{figure*}[t]
\begin{tabular}{cccc}
\includegraphics[width=0.22\linewidth]{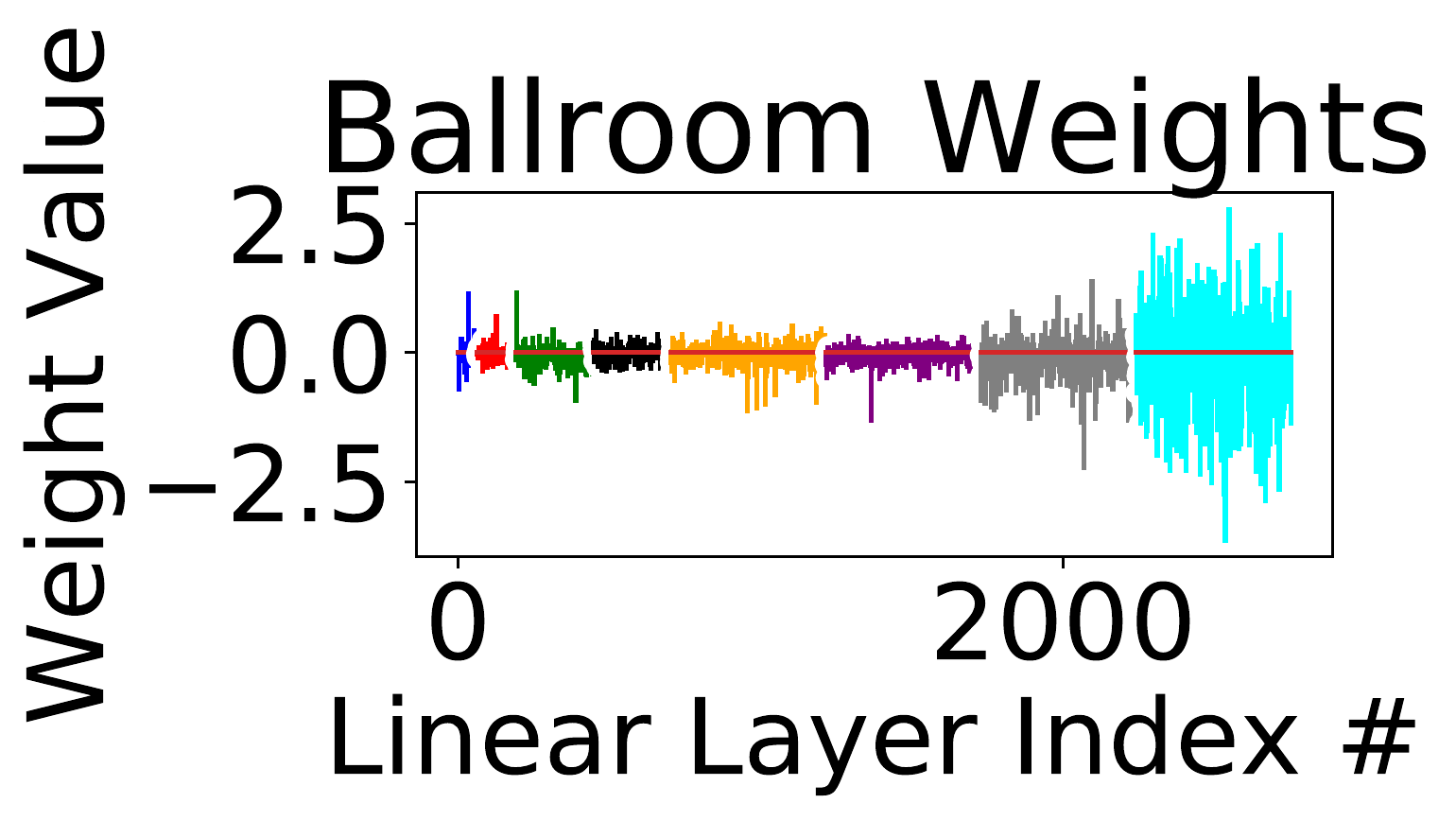}
&
\includegraphics[width=0.22\linewidth]{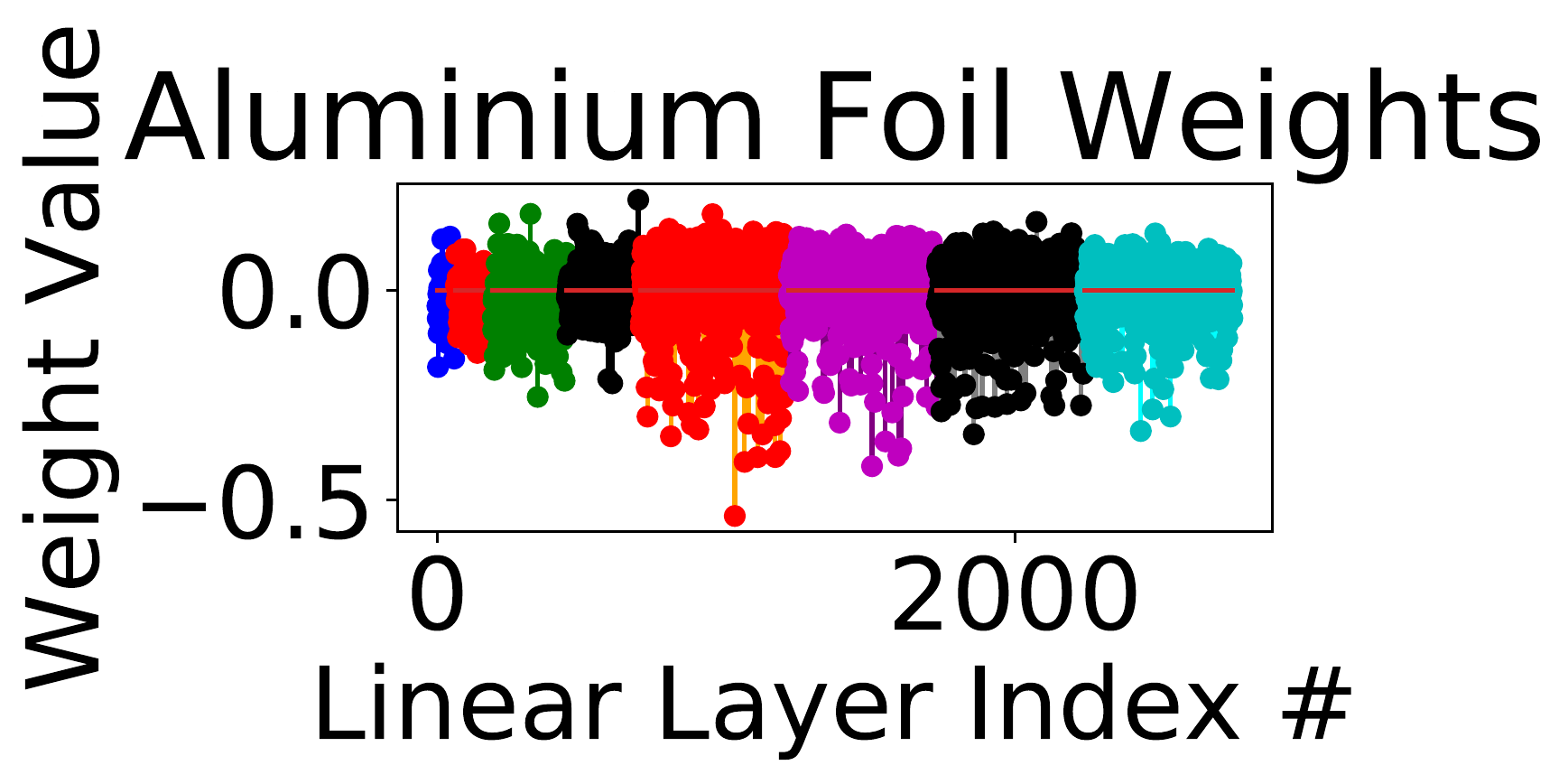}
&

\includegraphics[width=0.22\linewidth]{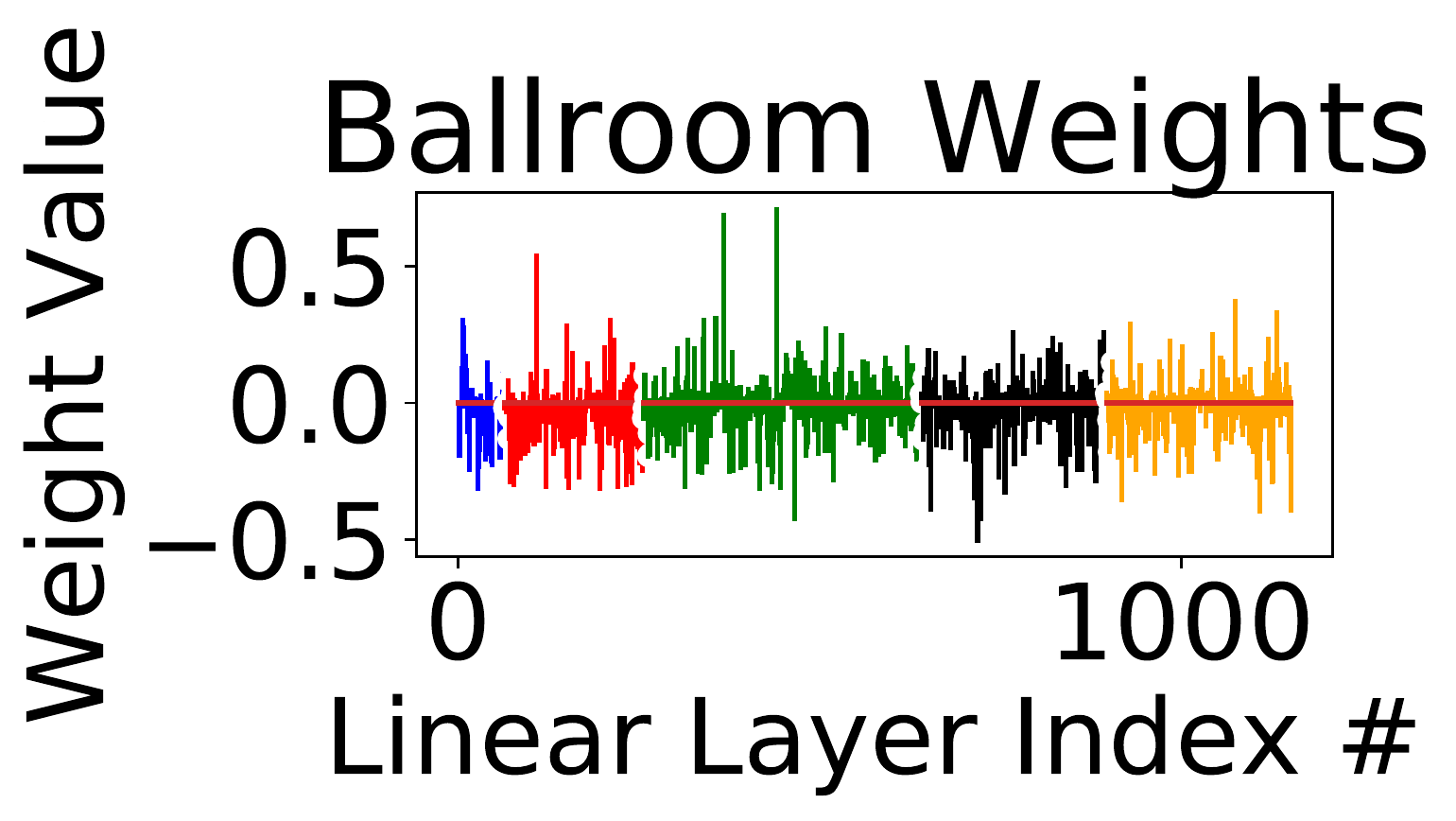}
&
\includegraphics[width=0.22\linewidth]{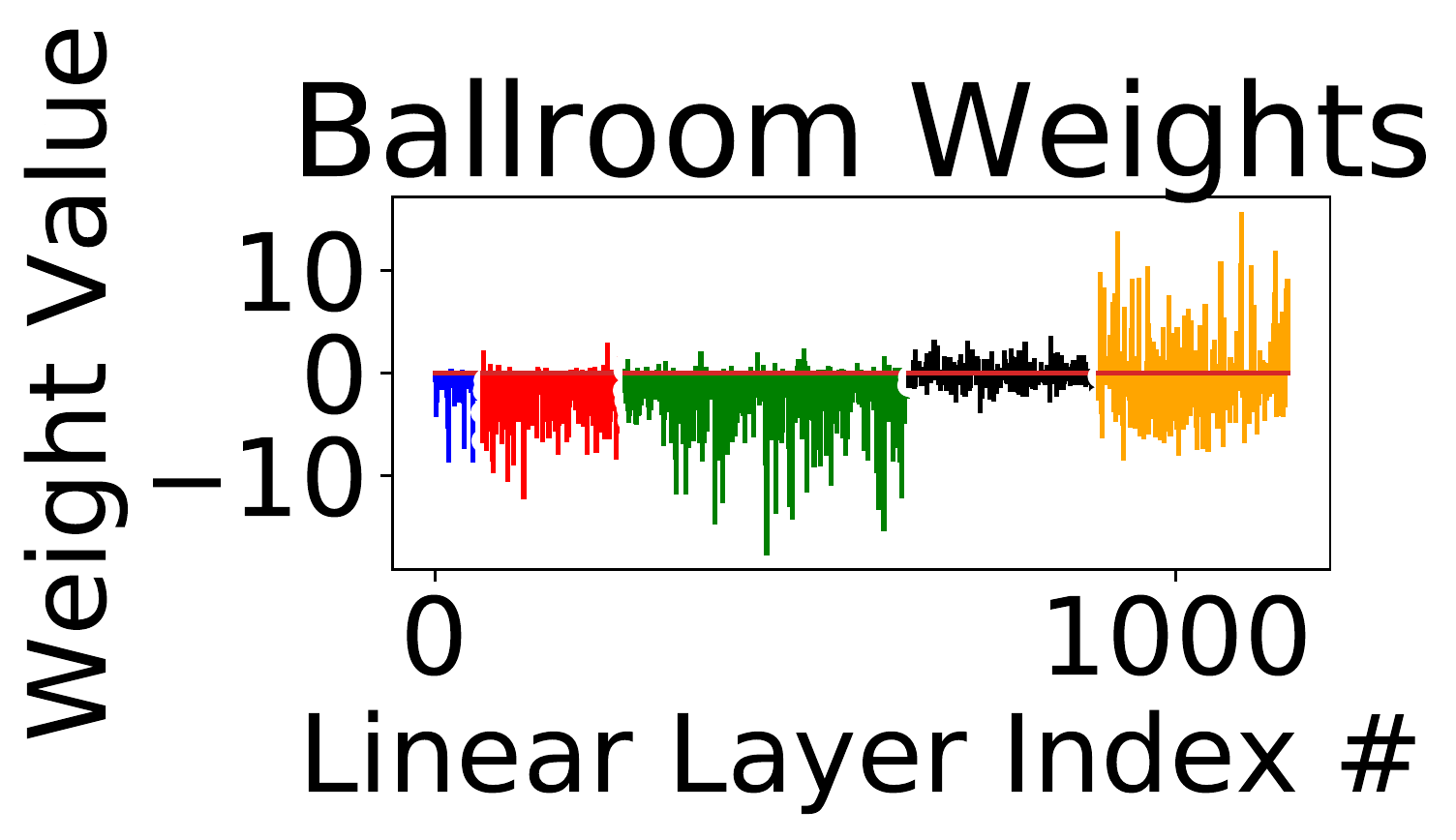}
\\
{\small (a) VGG11 Places10}
& {\small (b) VGG11 Texture}
& {\small (c) Alexnet Places10}
& {\small (d) Stylized Alexnet }
\end{tabular}
        \caption{
        \small 
        Weight values for varying NeuroView networks. (a) NeuroView VGG11, (c) Neuroview Alexnet, and (d) Stylized NeuroView Alexnet. While (b) is the weight values for the ``aluminum foil'' class of a NeuroView VGG11 network. The higher weights are closer to the deep layers for (a). In (b), the weight distribution is different from (a) and the main difference is the dataset for the network. In (c) for NeuroView Alexnet some of the higher weights are not associated with the last layer. In (d) is the same NeuroView Alexnet network but there are stylized images in the dataset. Additional training data changed how the network prioritizes the units. Each color represents a different convolutional layer.}
        
        \label{fig:BallroomLinearLayer}
        \vspace{-2ex}
\end{figure*}

Figures~\ref{fig:BallroomLinearLayer} (a) and (b) show the same NeuroView VGG11 network trained on two different datasets, Places 10 and Texture. Yet, the linear classifier has different prioritization with a dataset that has scenic classes while another dataset that has texture classes. Another benefit that NeuroView provides is to see the weights of the linear classifier to explain which units are contributing the most for that class.

\subsection{Case Study 3: Multi-View CNN}
 In \cite{su2015multi}, the authors construct a network where the input is 12 different camera rotated views of the same object. After the convolutional layers, the authors apply a view-pooling mechanism to acquire the maximum feature elements of the activations of all 12 views of the object. The issue is that the view-pooling mechanism may not be the most optimal approach and hence in this scenario we can use NeuroView to assess which views are the most important. Hence to convert the network, more will be modified compared to other NeuroView network conversions. For this scenario, we are using one NeuroView network for each camera rotated view and concatenating all the units' activations from each layer from each camera rotated view of the object. In this scenario, there were only 12 camera rotated views so then 12 CNNs were utilized, one for each view. From Table~\ref{tab:my_label}, the NeuroView network has better accuracy indicating that the view-pooling approach may not be the optimal choice in terms of accuracy.

\begin{figure*}[t]
\centering
\begin{tabular}{cc}
\includegraphics[width=0.48\linewidth]{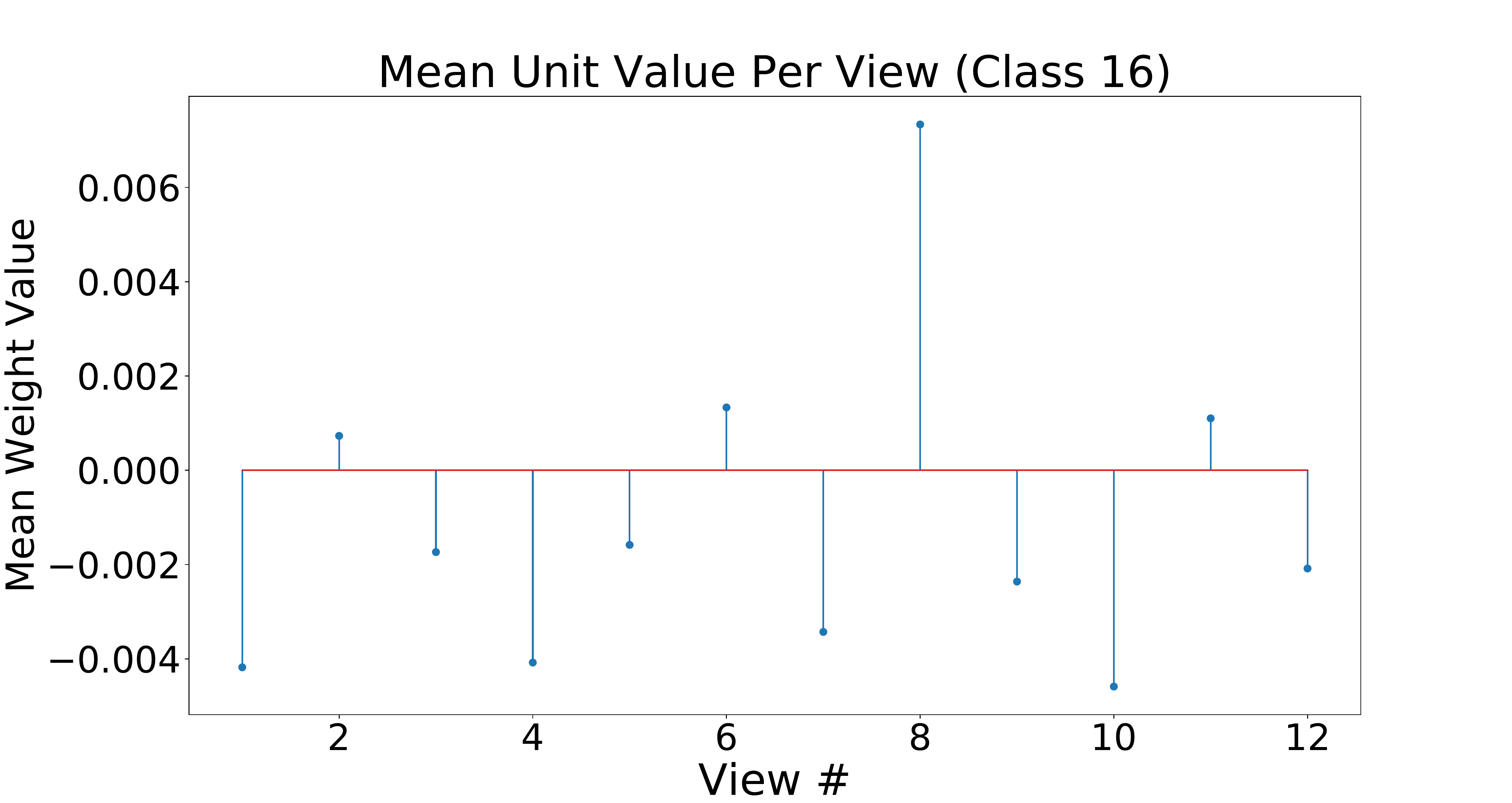}
&
\includegraphics[width=0.48\linewidth]{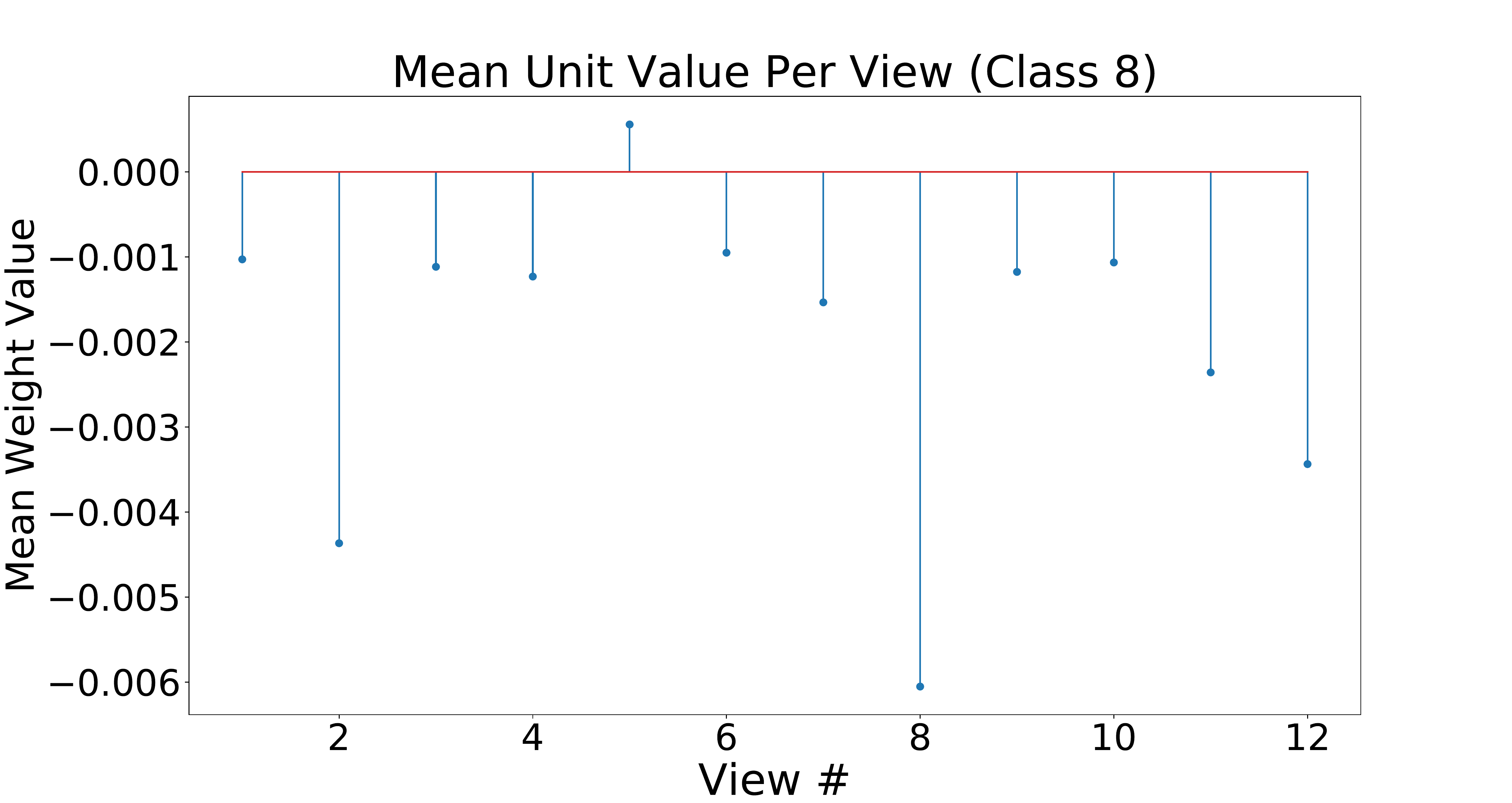}

\\
{\small (a) Class 16}
&
{\small (b) Class 8}

\end{tabular}
\caption{\small 
With NeuroView we can inspect the weights for the multi-view application to assess how the weights are being learned for the linear classifier. For each class we have the mean unit value for each view. The NeuroView weights show which view is important for the respective class. In class 16, there are four views that contribute positively. While in class 8, only one view contributes positively.}

\label{fig:MVWeights}
       \vspace{-2ex}
\end{figure*}

Figure~\ref{fig:MVWeights} shows the mean unit value for the views with the NeuroView network. The mean value is calculated by summing up all the weight values of the units for that view divided by the number of units. For each view, there are 2,752 units since we use the NeuroView VGG11 network. Figure~\ref{fig:MVWeights} (a) shows that for class 16, there are actually four views that are positive for the decision making. Hence NeuroView shows that there is at least one class where multiple views is beneficial to classification. In some cases, like class 8 (Fig.~\ref{fig:MVWeights} (b)), there is only one view where the mean weight for that view is positive. Even though, there are some classes with only one positive view weight, at least we verified using NeuroView. In that verification, we reveal that there are some classes where multiple views are beneficial. Thus, the view-pooling mechanism is not the most optimal approach and leveraging the other views' units led to a small performance in accuracy.

\subsection{Case Study 4: Sound Classification}
DNs have had a profound impact in computer vision tasks and from there people have adapted them to other domains. Particularly, researchers have been using CNNs that were originally used in computer vision. The issue is that one cannot see if the inner workings of the CNNs are the same with different tasks. Plus there are implications where relying on experimental findings is short sighted. For instance in \cite{recht2018cifar}, networks trained on CIFAR-10 did not generalize well to CIFAR-100. Similarly, in \cite{raghu2019transfusion}, the authors found that pretraining on natural images for medical imaging did not have any additional benefit. Hence, this is where NeuroView can provide the weights of the linear classifier to evaluate if for different tasks, the prioritization of the units will be the same or not.

For this scenario, the task is sound classification using a CNN network where the input are spectrograms, which are different from natural images which were used in the earlier applications. Prior methods from \cite{palanisamy2020rethinking, guzhov2021esresnet} and others that have used CNNs for the audio domain with spectrograms. Here we use an Alexnet and convert it to a NeuroView Alexnet network and in Table~\ref{tab:my_label} we have on par performance.

\begin{figure*}[t]
\centering
\begin{tabular}{cc}
\includegraphics[width=0.48\linewidth]{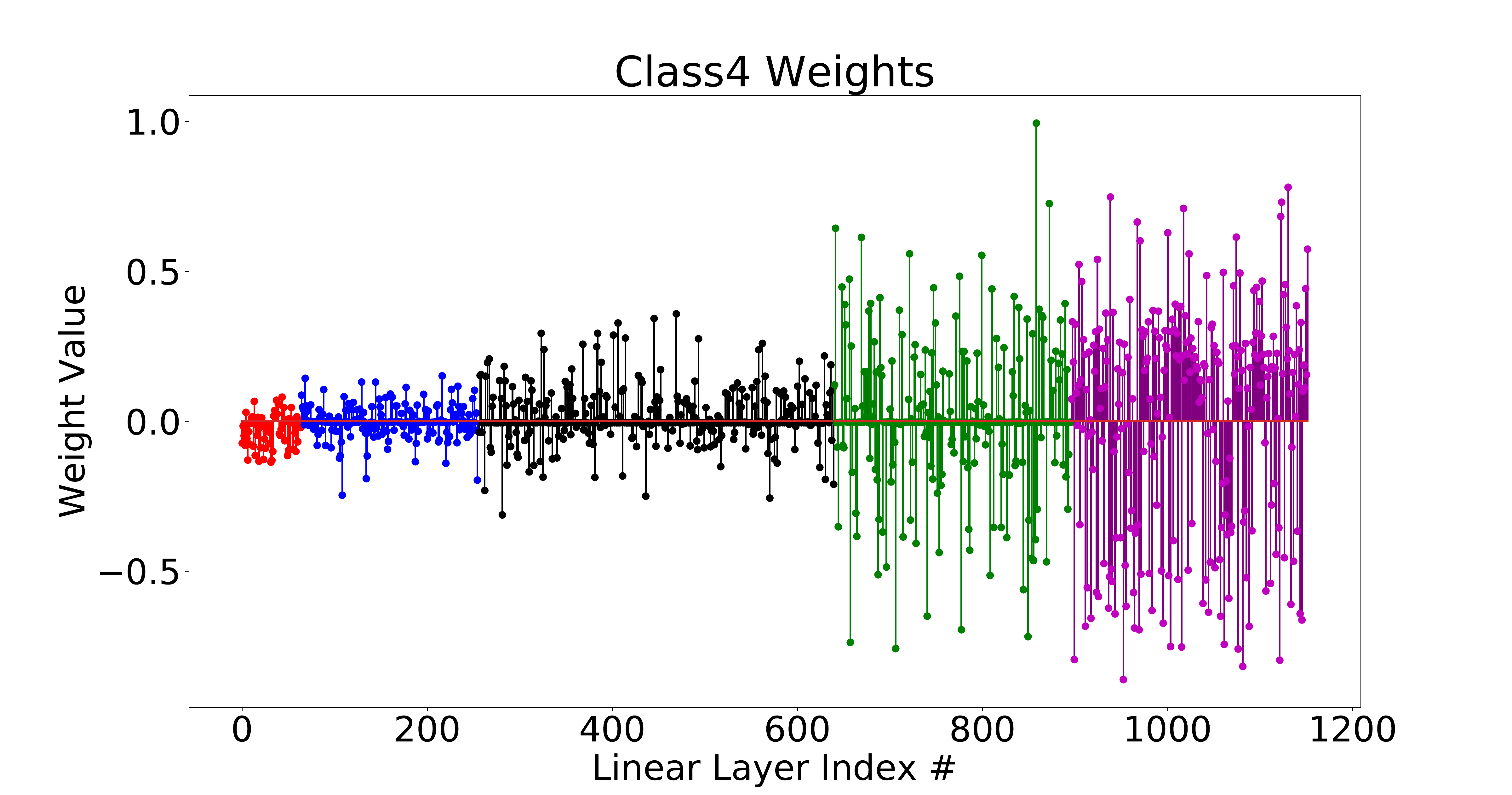}
&
\includegraphics[width=0.48\linewidth]{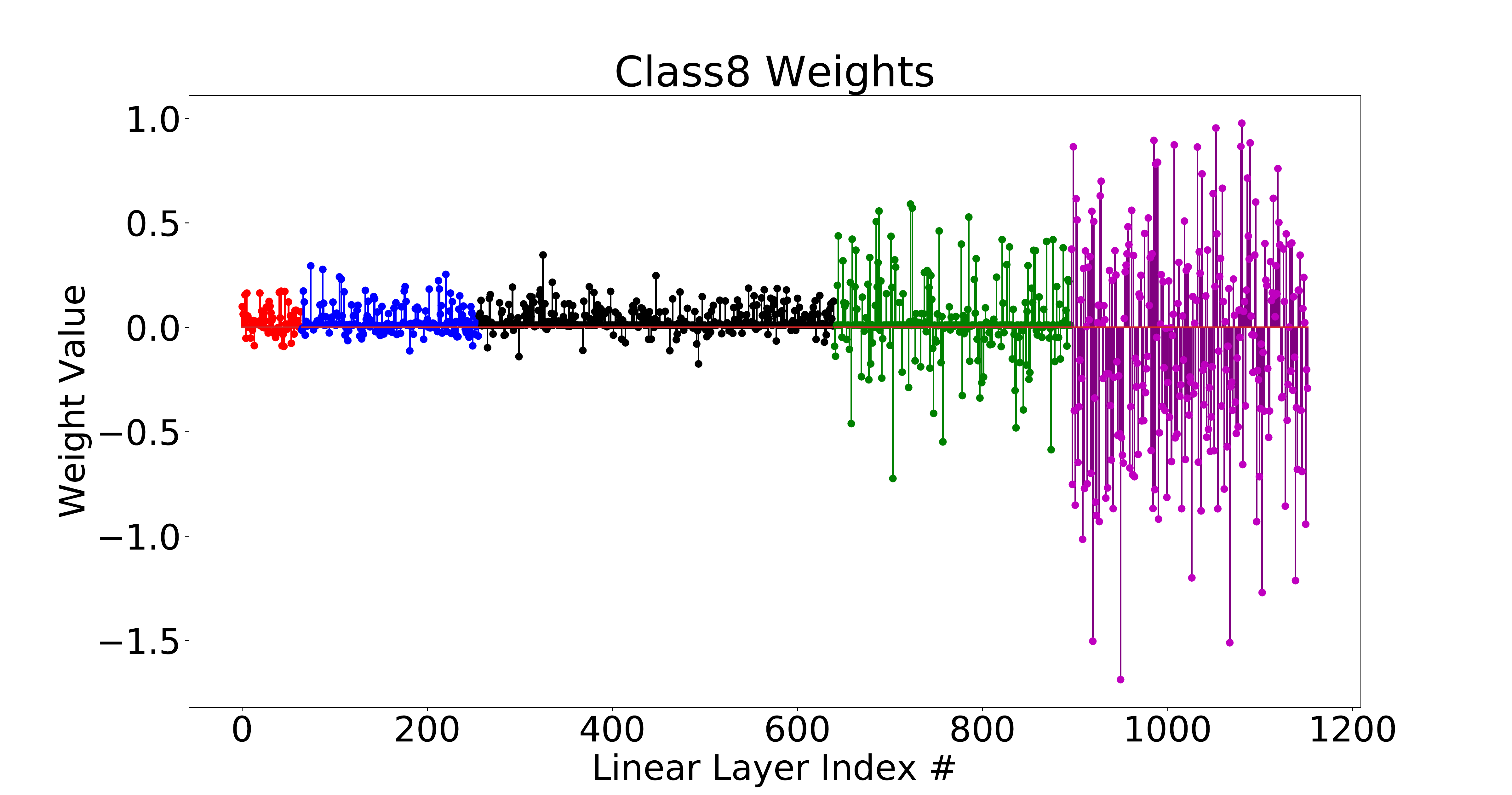}

\\
{\small (a) Class 4}
&
{\small (b) Class 8}

\end{tabular}
\caption{\small 
In using NeuroView we can inspect the weights in a dataset that uses spectrograms as the input for a CNN. We observe that for sound classification, the prioritization of units from the linear classifier is different from the same NeuroView network when trained on image datasets.}

\label{fig:AudioWeights}
       \vspace{-2ex}
\end{figure*}

In Figure~\ref{fig:AudioWeights}, we see that for two classes, class 4 and class 8, there are different mechanisms in play. For both class weight distributions, most of the significant positive weights were with the last two convolutional layers. This is interesting since with CNNs, the last layer's features are usually the input for linear classifier. Yet, here there is evidence that other layers' units are contributing to the decision-making process. One notable difference is that with class 8, there is a clear absence of negative weights for the first 3 convolutional layers compared to class 4 which has the weights of the first layer's units being negative. For different classes, there can be different mechanisms happening while the network is learning. Overall, by using NeuroView we observed a different behavior of the linear classifier's weights with this dataset which did not happen in Places10. Hence, it is important to understand the implications since if the behavior is different, then there could be pitfalls in the future.

\begin{figure*}[t]
\centering
\begin{tabular}{cccc}
\includegraphics[width=0.22\linewidth]{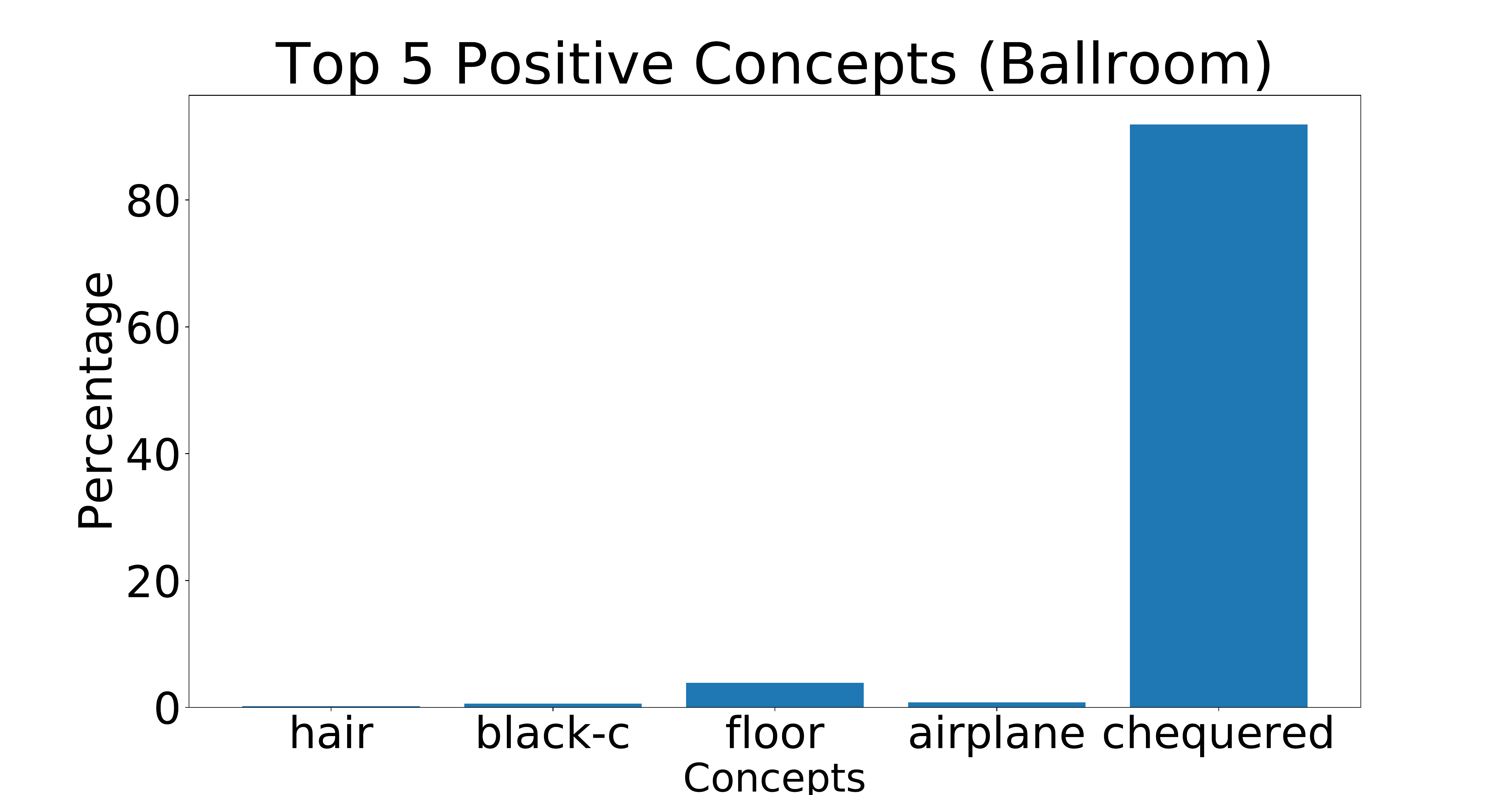}
&
\includegraphics[width=0.22\linewidth]{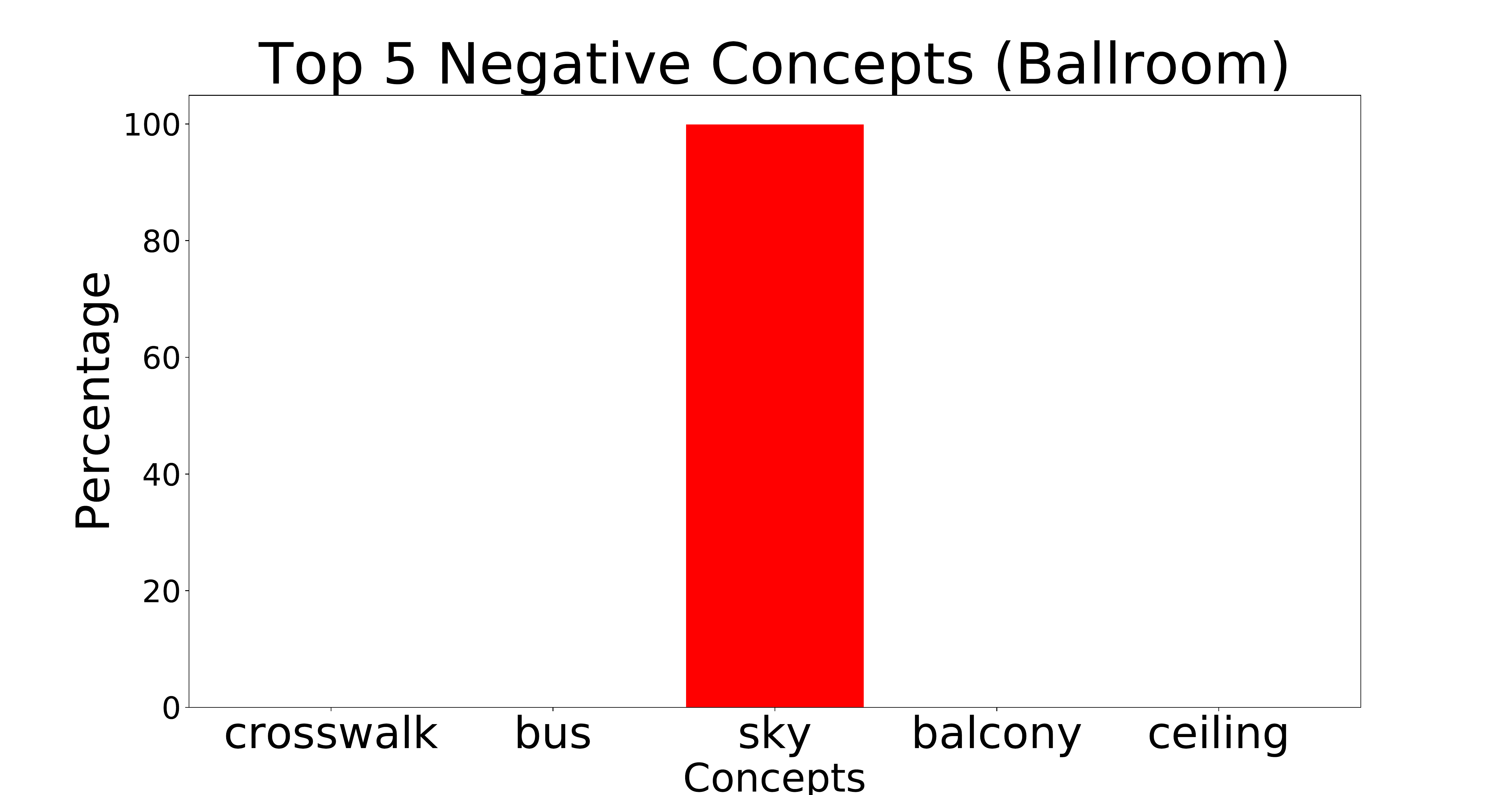}
&
\includegraphics[width=0.22\linewidth]{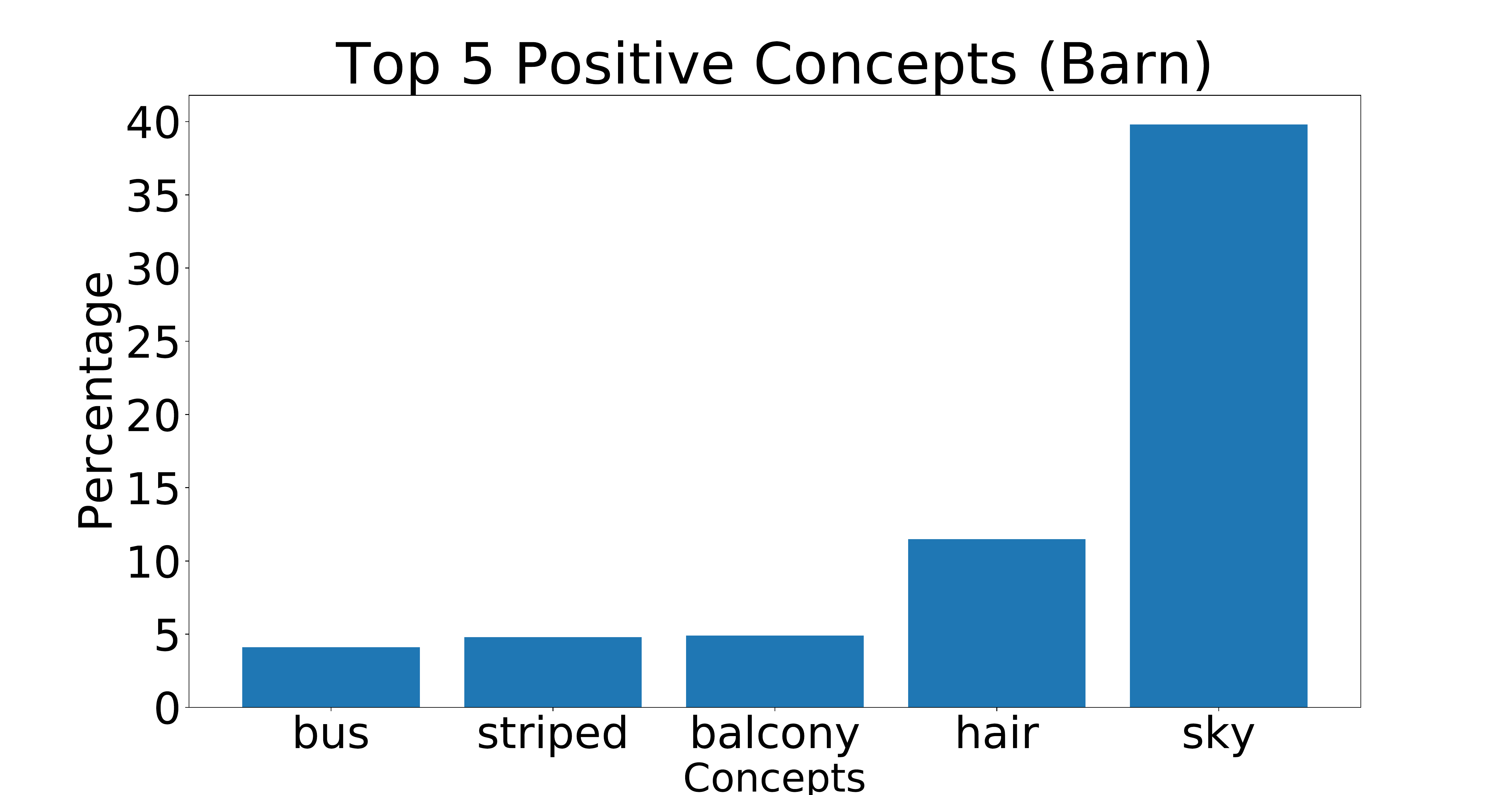}
&
\includegraphics[width=0.22\linewidth]{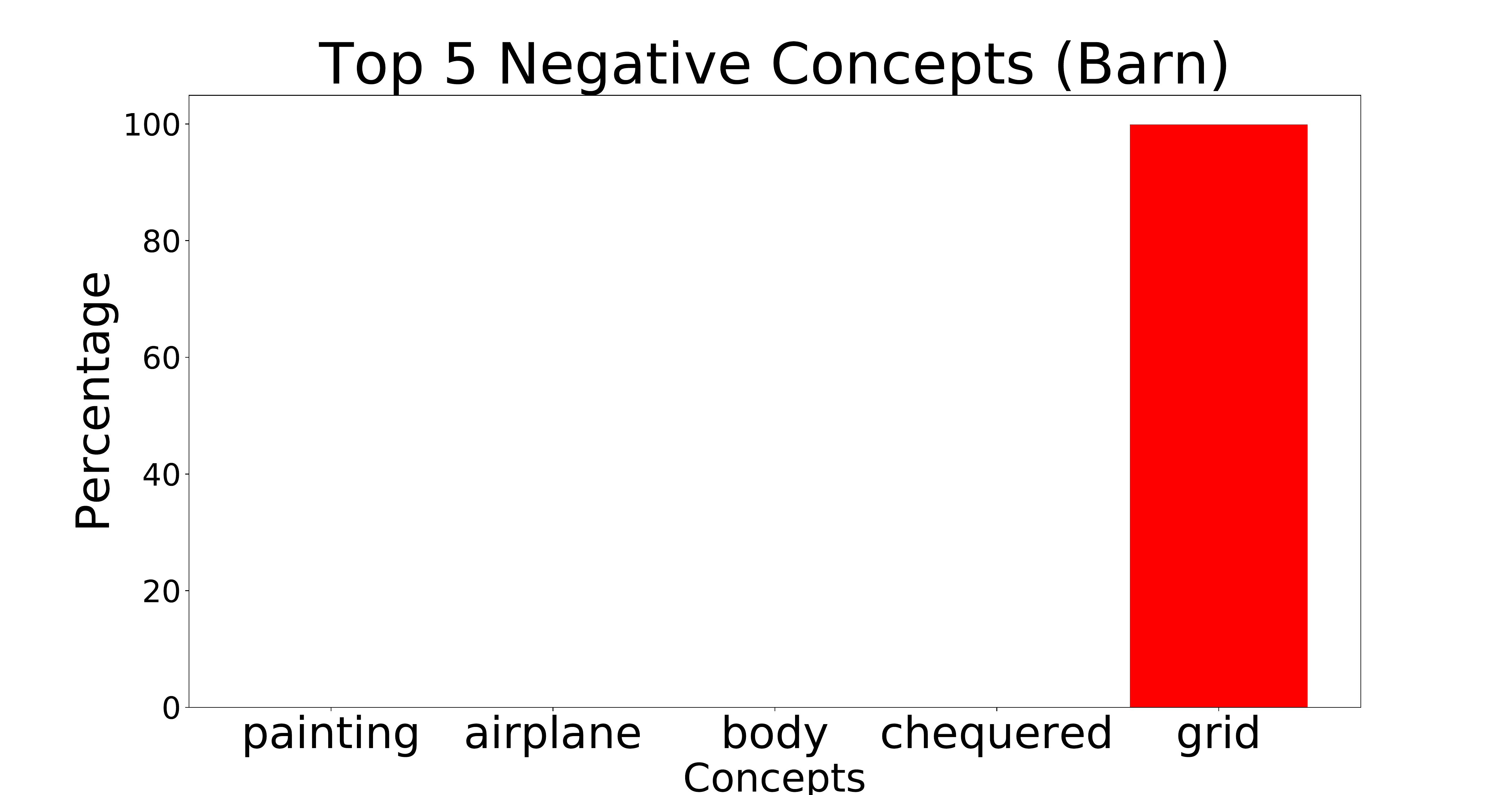}
\\
{\small (a) ``Ballroom'' +}
&
{\small (b) ``Ballroom'' --}
&
{\small (c) ``Barn'' +}
&
{\small (d) ``Barn'' --}
\end{tabular}
\caption{\small 
Using NeuroView and \cite{bau2017network}, each class can then be represented as set of concepts. Linking the concepts to each class helps provide more interpretability since each class has a set of concepts with percentages associated with them. For an NeuroView VGG11 network here are the top 5 concepts (positive (+) and negative (--)) for each class.}

\label{fig:ClassConcepts}
       \vspace{-2ex}
\end{figure*} 

\vspace*{-2mm}
\subsection{Case Study 5: Assessing Counterfactual Accuracy}
\begin{figure*}[t!]
        \centering
{\small (a)}
 \includegraphics[width=0.20\linewidth]{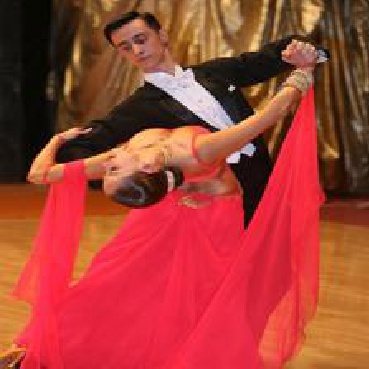}
 ~
{\small (b)}
  \includegraphics[width=0.20\linewidth]{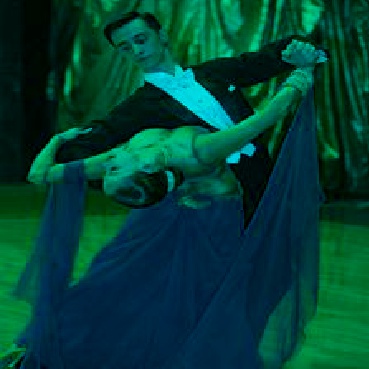}
  ~
{\small (c)}
  \includegraphics[width=0.20\linewidth]{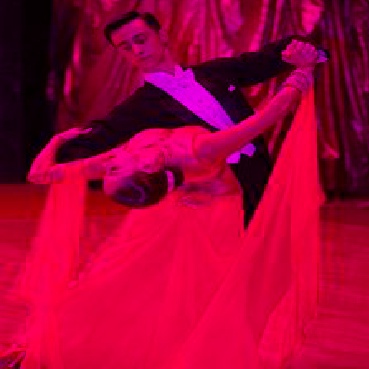}
  ~
{\small (d)}
  \includegraphics[width=0.20\linewidth]{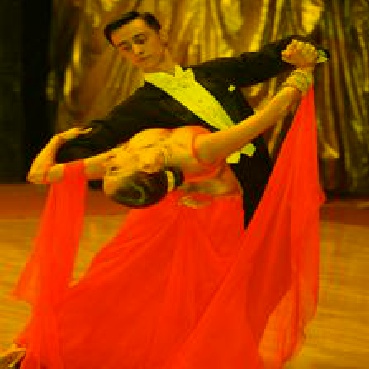}
        \caption
        {\small 
        \textbf{Color Perturbations:} ``Ballroom'' image with no modifications is the reference image (a). Red channel is omitted in (b), green channel is omitted in (c), and blue channel is omitted in (d). This is to assess how color can affect the classification accuracy.}
\label{fig:colorConcept}
\end{figure*}

Using the explainable NeuroView network we assess if perturbing certain aspects of the image will result in accuracy perturbation. Plus, in 4.1, we discovered which concepts mapped to certain classes. Hence, with the NeuroView networks we can investigate by perturbing the images to assess if the NeuroView networks' accuracy degradation will make sense. For instance, if an  arbitrary class had a set of particular colors in the dataset, then by perturbing those colors, the accuracy should drop. If an arbitrary class focuses more on texture than a color perturbation should not affect the performance as much.
For the validation set of the Places10 dataset, we set on of the color channels to zero to assess by how much the accuracy will drop. Table~\ref{tab:conceptAttacks} shows the validation accuracy among the different perturbation modifications. First off, in both NeuroView networks and DNs have a drop in accuracy. However, there is an interesting scenario that with the ``aquarium'' class with a NeuroView network had an increase in validation accuracy. This is an observation that is not observed with DNs and it does make sense for the ``aquarium'' class. The reason is that removing the red channel makes the images look more like a typical blue image. Figure~\ref{fig:colorConcept} (b) shows what a typical image with the red channel set to 0 looks. The interesting case is that setting the green or blue channel to 0 did not leave to a considerable drop in accuracy. This means that the NeuroView network is focusing more so on the texture than color for the  ``aquarium'' class.

In Table~\ref{tab:conceptAttacks}, there seem to be two different situations with the color perturbations with the NeuroView networks. In one situation, setting one color channel to 0 will drop the validation accuracy considerably. This happens with the ``barn'', ``baseball field'', and ``badlands'' classes. The second situation where the validation accuracy does not drop considerably like with the ``aquarium'' and ``ballroom'' classes. This seems to be that those two classes concentrate on the textures as opposed to the colors. In addition, ``aquarium'' and ``ballroom'' are more indoor settings compared to the scenic classes like ``barn''. In Figure~\ref{fig:ClassConcepts} (a), the dominant positive concept is a texture, chequered. Plus the most dominant negative concept for ``ballroom'' is sky so it also makes sense why when setting it to 0 will lead to a small increase in the validation accuracy since sky can contain some blue color.

\begin{table}[h]
\centering
\caption{Validation accuracy performance of NeuroView and DNs with different channel perturbations. Scenic classes like barn and baseball field lead to a greater drop of accuracy for NeuroView networks. While for indoor classes like ballroom, the color perturbations do not lead to a big drop of accuracy.}
\begin{tabular}{llllll}
\toprule
\multirow{3}{*}{\textbf{Network}} & \multirow{3}{*}{\textbf{Class}} & \multicolumn{4}{c}{\textbf{Validation Accuracy}} \\
    &   &       \multicolumn{4}{c}{Channel Perturbations}\\ 
  &   &  None & Red & Green & Blue \\
\midrule 
VGG11 & Barn & 95 & 0 & 71 & 8 \\ 
VGG11 & Aquarium & 88 & 84 & 18 & 71 \\
VGG11 & Ballroom & 88 & 32 & 45 & 75 \\
VGG11 & Baseball Field & 95 & 0 & 30 & 0 \\
VGG11 & Badlands & 95 & 0 & 0 & 20 \\ \hline
NeuroView VGG11 & Barn & 94 & 63 & 17 & 77 \\
NeuroView VGG11 & Aquarium & 91 & 98 & 85 & 89 \\
NeuroView VGG11 & Ballroom & 88 & 74 & 87 & 89 \\
NeuroView VGG11 & Baseball Field & 98 & 64 & 0 & 53 \\ 
NeuroView VGG11 & Badlands & 90 & 25 & 1 & 57 \\
\bottomrule
\end{tabular}
\label{tab:conceptAttacks}
\end{table}

  ~
        %

\vspace{-2ex}
\section{Conclusions}
With NeuroView, we can provide more analysis about how the units coordinate together for classification tasks. This type of explainability is important to understand the implications of what the units are learning and how they contribute to the decision making. With different case studies, we saw that the same NeuroView network prioritized on different units for different datasets which shows that it is important to understand what is happening within the network. NeuroView provides additional understanding and opens up more research questions. For instance, in other datasets like texture and spectrograms, the last convolutional layer is not the most prioritized as others have done before when using the convolutional layer's features as input to the fully-connected layer.



\newpage

\section*{Acknowledgement}
This work was supported by NSF grants CCF-1911094, IIS-1838177,
and IIS-1730574; ONR grants N00014-18-12571, N00014-20-1-2787, and
N00014-20-1-2534; AFOSR grant FA9550-18-1-0478; and a Vannevar Bush
Faculty Fellowship, ONR grant N00014-18-1-2047.
\bibliography{aaaa}
\bibliographystyle{plain}



\end{document}